\documentclass[letterpaper, 10 pt, conference]{ieeeconf} 
\usepackage{graphicx} 						
\usepackage{times}
\usepackage{amsmath} 					
\usepackage{amssymb} 						
\usepackage{subcaption}
\usepackage{amsmath,tikz}
\usepackage[export]{adjustbox}
\usepackage{footnote}
\usepackage{breqn}
\usepackage[colorlinks=true,citecolor=green,linkcolor=blue,urlcolor=blue]{hyperref}
\usepackage{hyperref} 
\usepackage{breakurl}
\usepackage{epstopdf}
\usepackage{tikz}
\usetikzlibrary{shapes,arrows, circuits}
\DeclareGraphicsExtensions{.eps}
\usepackage{schemabloc}
\usepackage{verbatim}
\pdfoutput=1
\hypersetup{linkbordercolor=green}

\tikzstyle{block} = [draw, fill=white!20, rectangle, 
minimum height=3em, minimum width=4em]
\tikzstyle{sum} = [draw, fill=white!20, circle, node distance=1cm]
\tikzstyle{input} = [coordinate]
\tikzstyle{output} = [coordinate]
\tikzstyle{pinstyle} = [pin edge={to-,thin,black}]

\IEEEoverridecommandlockouts

\title{\LARGE \textbf{
		Vision-based Control of a Soft Robot for Maskless Head and Neck Cancer Radiotherapy}
}

\author{Olalekan P. Ogunmolu$^{1}$, Xuejun Gu$^{2}$, Steve Jiang$^{2}$, and Nicholas R. Gans$^{1}$  % <-this % stops a space
	\thanks{$^{1}$Olalekan P. Ogunmolu and Nicholas R. Gans are with the Department of Electrical Engineering,
		University of Texas at Dallas, Richardson, TX 75080, USA
		{\tt\small \{olalekan.ogunmolu, ngans\}@utdallas.edu}}%
	\thanks{$^{2}$Xuejun Gu and Steve Jiang are with the Department of Radiation Oncology,  
		University of Texas Southwestern Medical Center, Dallas TX 75390, USA
		{\tt\small \{Xuejun.Gu, Steve.Jiang\}@utsouthwestern.edu}}%
}

\begin{document}
	
	%	\long\def\/*#1*/{}								% Define block comment
	
	\graphicspath{ {Charts/} }
	
	\maketitle
	\thispagestyle{empty}
	\pagestyle{empty}

	%%%%%%%%%%%%%%%%%%%%%%%%%%%%%%%%%%%%%%%%%%%%%%%%%%%%%%%%%%%%%%%%%%%%%%%%%%%%%%%%
	\begin{abstract}		
		This work presents an on-going investigation of the control of a pneumatic soft-robot actuator addressing accurate patient positioning systems in maskless head and neck cancer radiotherapy. We employ two RGB-D sensors in a sensor fusion scheme to better estimate a patient's head pitch motion. A system identification prediction error model is used to obtain a linear time invariant state space model. We then use the model to design a linear quadratic Gaussian feedback controller to manipulate the patient head position based on sensed head pitch motion. Experiments demonstrate the success of our approach.
			\iffalse
	\fi
	\end{abstract}

	%%%%%%%%%%%%%%%%%%%%%%%%%%%%%%%%%%%%%%%%%%%%%%%%%%%%%%%%%%%%%%%%%%%%%%%%%%%%%%%%
	\section{Introduction}
	This paper presents a continuation of our investigation of an image-guided soft robot patient positioning system for use in head and neck (H\&N) cancer radiotherapy (RT). % H\&N cancers are among the major cancers that cause a high patient death rate. 
	In 2014, over 1.6 million patients developed pharynx and oral cavity cancers in the United States, which led to over 580,000 deaths \cite{soft_robot:c1}.  	
	Typical H\&N cancer treatment involves intensity-modulated radiotherapy (IMRT), which delivers high potent dose to tumors while simultaneously minimizing dose to adjacent critical organs such as spinal cord, parotids glands, and optical nerves.  Typically, a patient lies on a 6-DOF movable treatment couch, and laser or image-guidance systems are used to ensure the patient is in the proper position. %A linear accelerator in conjunction with multileaf collimators adjusts a radiation beam to the shape of the patient's tumor. The beam is typically directed to a tumor location by accurately positioning the beam generator and/or moving the couch.	

IMRT requires accurate patient positioning. %In a geometric miss, for instance, highly conformal potent dose increases risk of underdose to tumors or undesirable high dose to critical organs and nearby tissues. 
An examination of patient displacement and beam angle misalignment during IMRT showed errors as small as 3-mm  caused 38\% decrease in minimum target dose or 41\% increase in the maximum spinal cord dose \cite{soft_robot:c2}. 	Image-guided radiotherapy (IGRT) has improved IMRT accuracy while reducing set-up times \cite{soft_robot:c7, soft_robot:c8, soft_robot:c6}. However, current IGRT practices focus on using images acquired before treatment to confirm beam placement \cite{soft_robot:Jaffray2012}. The discomfort caused by head masks in prolonged IMRT treatment can increase patients’ voluntary and involuntary motion. Studies show that translational errors caused by patient motion can be larger than 6mm, and rotational errors can be as high as 2$^{\circ}$ \cite{soft_robot:Kang2011}. Current motion-tracking systems, such as Cyberknife and Novalis are not compatible with conventional linear particle accelerators used at the majority of cancer centers.  Moreover, these two systems are limited to assuming the patient's body is rigid during motion tracking and compensation.	
Recently, a robotic real-time surface image-guided positioning system was studied for feasibility in frameless and maskless cranial stereotactic radiosurgery \cite{soft_robot:cervino}. While it achieved similar accuracy as the existing clinical methods, the system may not be suitable to IMRT due to the presence of mechanical and electrical parts in the path of the radiation beam.
	
	%The Novalis system can detect rotational set-up errors with an average accuracy of $0.09^{\circ}$ (standard deviation, $\sigma$, $0.06^{\circ}$), $0.02^{\circ}$ ($\sigma$, $0.07^{\circ}$) and $0.06^{\circ}$ ($\sigma$, $0.14^{\circ}$) for longitudinal, lateral and vertical rotations. It has an average positioning accuracy of $0.06^{\circ}$ ($\sigma$, $0.04^{\circ}$), $0.08^{\circ}$ ($\sigma$, $0.06^{\circ}$) and $0.08^{\circ}$ ($\sigma$, $0.07^{\circ}$) for longitudinal, lateral and vertical rotations respectively \cite{soft_robot:RTM}. 
	
%Recently, a 6D robotic real-time surface image-guided positioning system was employed in a feasibility study for frameless and maskless cranial stereotactic radiosurgery (SRS) \cite{soft_robot:cervino}. It tested the accuracy of a surface-image guided procedure against an optical guidance platform (OGP) used to treat SRS at many institutions.  This system is designed to be more comfortable than traditional systems, and achieved similar accuracy as the OGP.  However, the procedure emphasized visual inspection and cooperation of the patient to assure immobilization during therapy. This may be impractical, since it assumes patient cooperation and requires treatment be interrupted whenever patient motion is beyond pre-defined tolerance.

Soft robot systems are deformable polymer enclosures with fluid-filled chambers that enable manipulation and locomotion tasks  by a proportional control of the amount of fluid in the chamber \cite{soft_robot:soft1, soft_robot:soft2}. Their customizable, deformable nature and compliance make them suitable to biomedical applications as opposed to rigid and stiff mechanical robot components.  They can also be made radiotransparent, which is necessary in IMRT.	
	
The long term goal of our work is to address the non-rigid motion compensation during H\&N RT. As we continue our initial investigation, we control one degree of freedom, raising or lowering of a generic patient's head, lying in a supine position, to a desired height above a table. The current system consists of a single inflatable air bladder (IAB), a mannequin head and a neck/torso motion simulator, 
two different Kinect RGB-D cameras to measure patient position, 
two current-controlled pneumatic valve actuators, and a National Instruments myRIO microcontroller. In this work, we extended and improve our previous work  \cite{soft_robot:paper1}. This paper contributes better vision tracking and localization methods via filtering and fusion of the two RGB-D estimates. We improve on the system identification of the soft-robot system and now incorporate an optimal control network.  The result is a much improved motion control.  
		
Section \ref{sec:hardware} of this paper briefly presents the design of the soft robot system.  Section \ref{sec:Vision} discusses the computer vision algorithms to 
detect the patient's face and fusion of measurements from the RB-D images.  Section \ref{sec:sys_ID} presents results of system identification for the soft-robot system.  Section \ref{sec:LQG} presents design of the linear quadratic Gaussian (LQG) controller, and Section \ref{sec:exps} presents several experiments to demonstrate
the system.
		
	\section{Soft Robot Design Overview}\label{sec:hardware}
	The soft robot actuation mechanism combines an IAB (19" x 12") made of lightweight, durable and deformable polyester and PVC, two current-controlled proportional solenoid valves, and a pair of silicone rubber tubes (attached to a T-port connector at the orifice of the IAB) in order to convey air in/out of the IAB. A 1HP air compressor supplied regulated air at 30 psi to the inlet actuating valve, while an interconnection of a 60W micro-diaphragm pump and a valve removed air from the outlet terminal of the IAB. 	The RGB-D sensors are mounted directly above the head for raw head position and velocity measurements, while local Kalman filters (KFs) provide two estimates of the head position and velocity. The sensor estimates are aggregated using a track-to-track KF-based sensor fusion algorithm. 	
	We apply the fusion result in a new robust control law for the pneumatic actuator valves, thereby regulating air pressure within the IAB and moving the patient's head as desired. The real-time controller was deployed on a National Instruments myRIO embedded system running LabVIEW 2015. The LabVIEW algorithms were processed within a Windows 7 virtualbox running on the Ubuntu host workstation. 
	%The previous system set-up is described in details in \cite[\S 1]{soft_robot:paper1} 
%	Soft robot systems are typically constructed from deformable polymer enclosures with fluid-filled chambers, and they do enable manipulation and locomotion tasks through the control of the amount of fluid in the enclosed chamber \cite{soft_robot:soft1, soft_robot:soft2}. Their customizable, deformable nature and compliance make them suitable to biomedical applications as opposed to rigid and stiff mechanical robot components -- impractical in enabling articulation of human body parts. 
	
%	\section{System Overview}\label{sec:sys_des}%
We use a Kinect Xbox 360, and a Kinect for Windows v2 sensor to estimate head position and velocity. The two sensors use different electronic perception technologies to determine distance of an object from the camera origin. They therefore have different lateral and range resolutions as well as different noise characteristics. Image processing for both cameras is executed on a 22GB RAM mobile workstation with Intel Core i7-4800MQ processor running 64-bit Ubuntu Trusty on a Linux 4.04 kernel. 
%	\begin{figure}[tb]
%		\centering
%		\includegraphics[keepaspectratio = false, width=3.6in, clip=true, angle=0, center]{Expt_Environment_2.pdf}
%		\vspace{0.1em}
%		\caption{\footnotesize Experimental Testbed}
%		\label{Fig. 1.}
%	\end{figure}
%	\vspace{-0.3em}	
	\section{Image-Based Patient Position Estimation}   \label{sec:Vision}	
	We perform recursive filter estimations of the RGB-D measurements and improve position estimates by using an additional sensor to better localize tracked features. 
		\begin{figure}[tb]
			\centering
			\includegraphics[trim=1cm 0cm 3cm 0cm, width=3.5in, clip=true, angle=0, center]{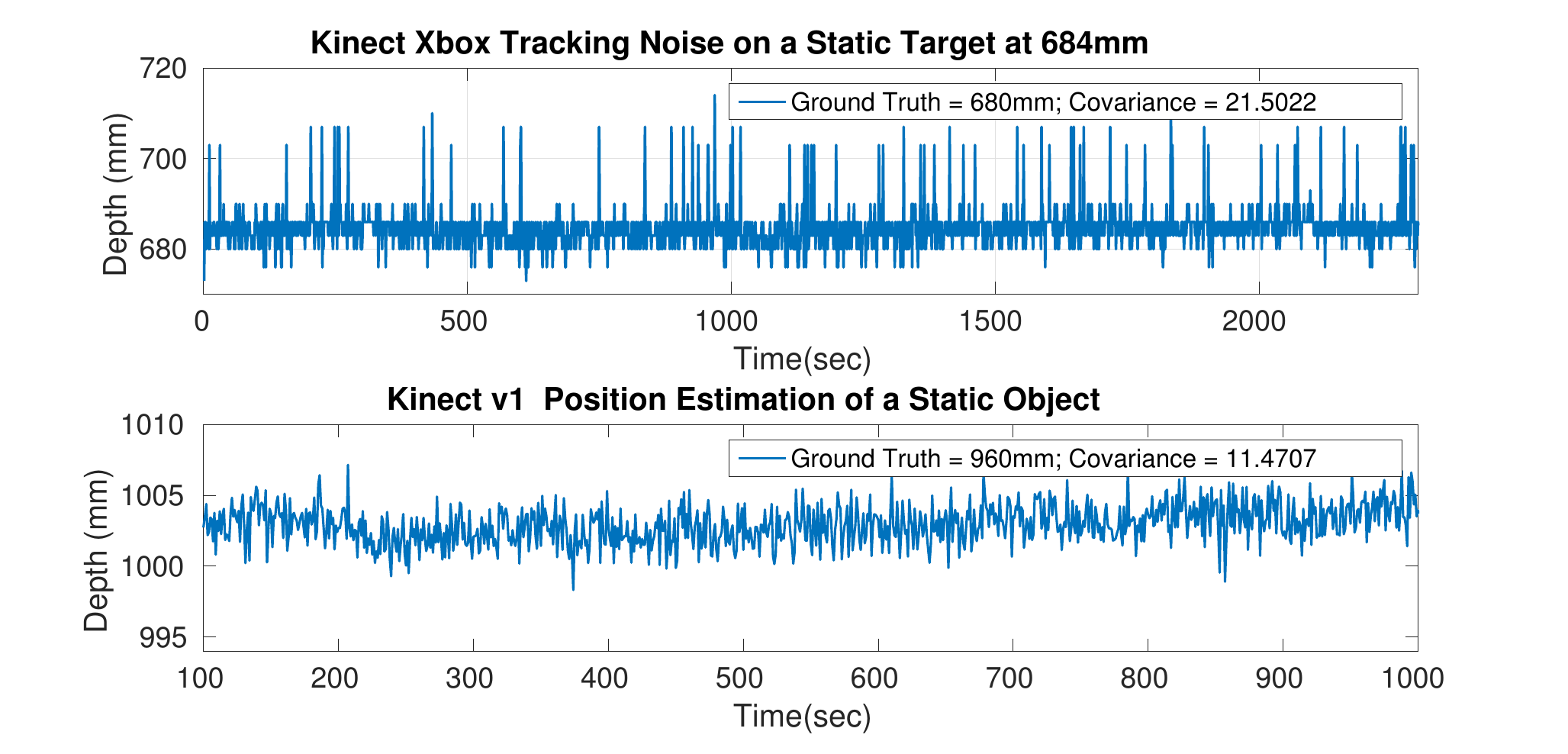}
			%\vspace{0.1em}
			\caption{Noise floor of Kinect Xbox Sensor vs. Kinect v2 Sensor}
			\label{fig:cam_noises}
			\vspace{-0.9em}
		\end{figure}
	We add the Kinect v2 sensor (henceforth called the v2 sensor), based on the time-of-flight (ToF) electronic perception principle. In ToF, light pulses illuminate a scene, and depth is calculated by determining the phase shift of the returned light signals. The active infra-red reduces the dependence on ambient lighting \cite{soft_robot:Kinectv2Features}, and this sensor has a higher spatial depth resolution of 512$\times$424 pixels at 30Hz interactive rate, compared to the Xbox's 320 $\times$ 240 pixels \cite{soft_robot:ms_constants}. To minimize the noise due to the limited sensor resolution, the v2 has in-built noise improvement capabilities \cite{soft_robot:canesta}.

	The v2 provides a higher depth-map accuracy and lower noise floor compared to Kinect for Xbox, as can be seen from \autoref{fig:cam_noises}, where the v2 exhibits a noise auto-covariance of 11.4707$mm^2$ compared with  22.7057$mm^2$ for the Xbox.	Despite the improved performance of the v2, noise remains an issue, as is the case for every electronic perception system. To alleviate this, we employ a multisensor data fusion of both Kinect sensors' observations. We achieved this by local Kalman Filter estimates of each sensor's observations, and we fuse the estimates via a variance-weighted multisensor Kalman filter fusion scheme described later in this section.

	\subsection{Face Detection and eye-feature tracking}
	We approached face detection using Haar Cascade Classifiers (HCC) \cite{soft_robot:HaarDetectors}. HCC's  are based on integral image representation, which allow for features evaluation while maintaining high detection rates. The features resemble Haar basis functions.  A classifier is formed by choosing a small number of crucial features with AdaBoost, and a weighted sum of individual classifiers is used to construct a strong object detection classifier in a cascade manner. This \iffalse minimizes having to search through all pixels in an image when looking for an object of interest and \fi increases the detector's speed by concentrating on areas within an image with high probability of features of interest.	
	\begin{figure}[tb]
		\centering
		\includegraphics[ width=3.2in,  clip=true]{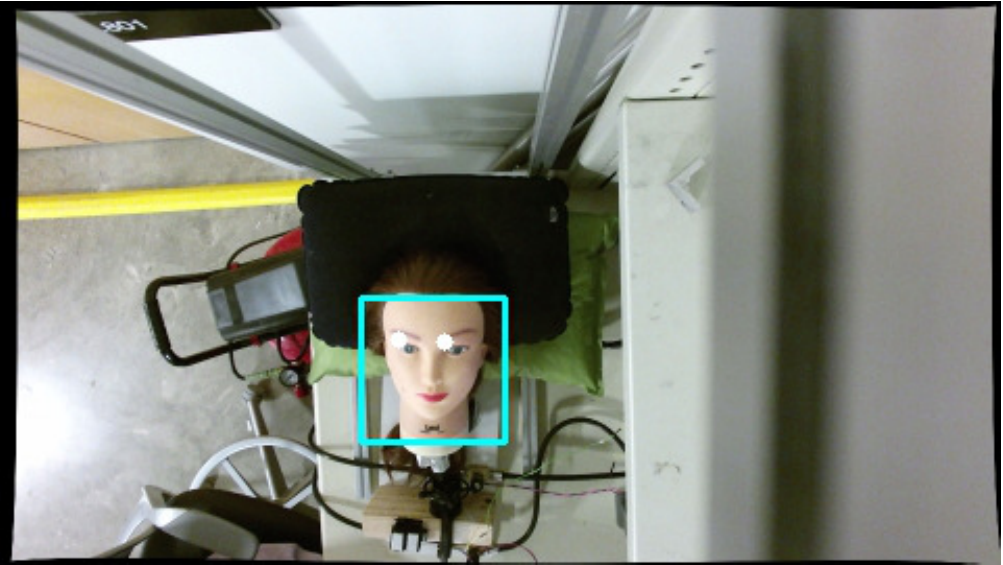}
	%	\hspace{0.05cm}
	%	\includegraphics[ width=1.6in,  clip=true]{Charts/Face_Gray2.jpg}
		\vspace{0.7em}
		\caption{Original colored image retrieved from the Kinect v2 Sensor. 
		%	Right: A down-sampled gray scale image used for the face/eye detection.
		}
		\label{fig:down}
		\vspace{-0.4em}
	\end{figure}
	
	%We used the OpenCV-trained Haar frontalface face and eye classifiers.
	A drawback of HCC's is the memory consumed on computing devices when searching through image pixels for specific regions of interest. Searching through a 640 $\times$ 480 pixels gray-scale image for specific features caused a 90\% reduction in the frame rates of either sensor, when the algorithm is run on a CPU. To overcome this, both sensor's images were spatially down-sampled via linear interpolation before HCCs were applied. Face detection was performed on a single NVIDIA Quadro K1100M GPU. We retrieve each detected face from the GPU, and then detect eyes within detected faces using the same procedure. %See \autoref{fig:flow} for the detection procedure. 
	
To achieve robust detection, the minimum  number of neighbors in each candidate rectangle feature was determined based on our experience. The search area within an image was chosen to be within the range of (5 $\times$ 5) pixels and (20 $\times$ 20) pixels. This gave us more than 90\% face detection rate for both sensors.  A similar approach was used for the eye classifier. The final implementation achieved a frame rate of 15Hz for each sensor running independently on the Linux host computer. Further improvement in frame rates is an avenue for future work. %When run simultaneously, due to the heavy load on the CPU bus and USB controller, we achieved 11Hz frame rate after increasing the kernel USBFS memory buffer of the host OS by appending the LInux command line utility in grub. 
	
	%The expensive phase of the detection mechanism being moved to the GPU enabled us to run the detection algorithm at a frame rate of 15Hz for each sensor running simultaneously on the Linux host computer. To accommodate both  vision processing running in tandem on the same workstation, we increased the kernel USBFS memory buffer of the host OS by appending the Linux command line utility in grub. The result of the retrieved depth point has been shown earlier in \autoref{Fig. 2.} where we see that the Xbox camera's observation has 86\% more noise than the Kinect v2 sensor going by the estimated auto-covariance of each sensor's observation. Neither the observation of either sensor is good enough for our control requirement. To refine each observations, we compute local Kalman filter estimates for each sensor and use a variance-weighted multisensor Kalman filter fusion approach in obtaining less noisier signals.
\iffalse
	\begin{figure}[tb]
		\centering
		\includegraphics[width=3.4in, keepaspectratio=false, clip=true, center]{Flowchart.pdf}
		\vspace{0.1em}
		\caption{Flow-chart of Haar Face and Eye Detection}
		\label{fig:flow}
	\end{figure}
\fi	
	\subsection{Local Kalman Filters} \label{sec:KF}
	From \autoref{fig:cam_noises}, we see that both RGB-D sensors suffer from notable associated noise, which is not suitable for our control requirement. To refine the observation, local Kalman Filter (KF) estimates for each sensor were computed \iffalse \cite{simon2006optimal} \fi to determine state estimates $\hat{\textbf{x}}(i)$ that minimizes the mean-squared  error to the true state $\textbf{x}(i)$, given a measurement sequence
	$z(1), \cdots, z(j)$, that is
	\begin{align} \label{eq.est_expt}
	\hat{\textbf{x}}(i|j)&= \text{arg } \min_{\hat{\textbf{x}}(i|j)\in \mathbb{R}^n}
	\mathbb{E}\{(\textbf{x}(i) - \hat{\textbf{x}})(\textbf{x}(i) - \hat{\textbf{x}})|z(1), \cdots, z(j)\} 	\nonumber \\
	&\triangleq\mathbb{E}\{\textbf{x}(i)|{z}(1), \cdots, {z}(j)\} \triangleq\mathbb{E}\{\textbf{x}(i)|{Z}^j\}
	\end{align}
	where the obtained estimate is the expected value of the state at time $i$ given observations up to time $j$.	The covariance of the estimation error is given by
	\begin{align}
	\textbf{P}(i|j)\triangleq \mathbb{E}\{(\textbf{x}(i) - \hat{\textbf{x}}(i|j)(\textbf{x}(i) - \hat{\textbf{x}}(i|j)^T | Z^j\}.
	\end{align}
	
	Assuming the model of the state is common to both sensors, and denoting the distance from the v2 to the head as $d(k)$, we define $\textbf{x}(k)=[d(k),\,\dot{d}(k)]^T \in \mathbb{R}^2$ as the state vector of interest, and let $\Delta T $ be the time between steps $k-1$ and $k$. The model state update equations are given by %the Linear difference equation 
	\begin{equation}
	\textbf{x}_k = \textbf{F}_k\textbf{x}_{k-1}+\textbf{B}_k\textbf{u}_k+\textbf{G}_k\textbf{w}_k
	\label{eq:state_model}
	\end{equation}
	where 
	$\textbf{F}(k) \in \mathbb{R}^{2\times 2}$ is the state transition matrix given by
	\begin{equation}
	\textbf{F} = \begin{bmatrix}
	1 & \Delta T \\
	0	& 1
	\end{bmatrix} 
	\end{equation}
	$\textbf{u}(k) \in \mathbb{R}^2$ is the control input,
	$\textbf{B}(k)$ is the control input matrix that maps inputs to system states,
	$\textbf{G}(k) \in \mathbb{R}^{ 2 \times 2}$ process noise matrix, and
	$\textbf{w}(k) \in \mathbb{R}^2$ is a random variable that models the state uncertainty.
	In the absence of inputs $\textbf{B}_k\textbf{u}_k = 0$, and the
	model becomes
	\begin{align}\label{eq.accelmodel}
	\textbf{x}_k =  \textbf{F}_k \textbf{x}_{k-1}+ \textbf{G}_k \textbf{w}_k
	\end{align}
	where 
	$ \textbf{w}_k$ is the effect of an unknown input and $ \textbf{G}_k$ applies that effect to the state vector, $ \textbf{x}_k$.
	The process noise is assumed unknown and is modeled as uncontrolled forces causing an acceleration $a_k$ in the head position ($a_k$ is thus a scalar random variable with normal distribution, zero mean and standard deviation $\sigma_a$).  We model this into \eqref{eq:state_model} by setting $\textbf{G}_k$ to identity and set $\textbf{w}(k) \sim \mathcal{N}(0, \textbf{Q}(k))$  where the covariance matrix $\textbf{Q}(k)$ is set to a random walk sequence defined by
	$\textbf{W}_k={[\frac{{\Delta T}^2}{2}, \Delta T ]}^T$ . Therefore, we find that
	\begin{align}
	\textbf{Q} &= \textbf{W}\textbf{W}^T{\sigma_a}^2
	= \begin{bmatrix}
	\dfrac{{\Delta T}^4}{4} &	\dfrac{{\Delta T}^3}{2} \\
	\dfrac{{\Delta T}^3}{2} & {\Delta T}^2
	\end{bmatrix}{\sigma_a}^2.
	\end{align}
	
	Denoting the head displacement at time $k$ as measured by the Xbox and v2 as 
	${z}_1(k)$ and ${z}_2(k)$ respectively, the sensors' measurements were mapped to the v2 reference frame and modeled as 	
	\begin{equation}\label{eq:sensors_obs}
	{z}_s= \textbf{H}_s(k)\textbf{x}(k)+{v}_s(k) \qquad \qquad s = 1,2
	\end{equation}
	where $\textbf{H}_s(k) ={\begin{bmatrix}  1 & 0 \end{bmatrix} }^T$ maps the system's state space into the observed space, and ${v}_s(k) \in \mathbb{R}$ is a random variable that models the sensor error. We define ${v}_s(k)$ as a normally distributed random variable with zero mean and variance
	$\sigma_{rs}^2$.  We assume the random sequences ${v}_1(k)$,${v}_2(k)$, $\textbf{w}(k$) are independent and uncorrelated in time.
	\iffalse
	%\subsubsection{Local KF Implementation}
	The estimates that minimize the mean-squared error of either sensor's observation sequence $Z^j$ are initialized as 
	\begin{gather}
	\hat{\textbf{x}}_0 = \mathbb{E}(\textbf{x}_0) \nonumber \\
	\textbf{P}_0 = \mathbb{E}[(\textbf{x}_0 - \hat{\textbf{x}}_0){(\textbf{x}_0 - \hat{\textbf{x}}_0)^T}]	
	\end{gather}
	where $\textbf{P}_0$ is the associated covariance matrix of the estimate.
	\fi
	
	At each time step, $k$, each local KF's priori and posteriori estimates are computed through the typical prediction and update phases
	
	\textbf{Prediction Phase:}
	\begin{align}\label{eq:predict}
	\hat{\textbf{x}}_{k|k-1}&=\textbf{F}\hat{\textbf{x}}_{k-1|k-1} + \textbf{B}_k\textbf{u}_k  \nonumber \\ 
	\textbf{P}_{k|k-1}&=\textbf{F}_k\textbf{P}_{k-1|k-1}{\textbf{F}_k}^T + \textbf{Q}_k
	\end{align}
	where $\hat{ \textbf{x}}_{k|k-1}$ and $ \textbf{P}_{k|k-1}$ are the state prediction vector and the prediction covariance matrix respectively.
	
	\textbf{Update Phase:}
	\begin{align} \label{eq:update}
	\textbf{K}_k &=  \textbf{P}_{k|k-1}{ \textbf{H}_k}^T{[ \textbf{H}_k \textbf{P}_{k|k-1}{ \textbf{H}_k}^T+ \textbf{R}_k]}^{-1}
	\nonumber \\ 
	\hat{ \textbf{x}}_{k|k}&=\hat{ \textbf{x}}_{k|k-1} +  \textbf{K}_k ( \textbf{z}_k -  \textbf{H}_k \hat{ \textbf{x}}_{k|k-1}) %\tilde{y}_k  
	\nonumber \\ 
	\textbf{P}_{k|k}&=( \textbf{I} -  \textbf{K}_k \textbf{H}_k) \textbf{P}_{k|k-1}
	\end{align}
	where $ \textbf{K}_k$, $\hat{ \textbf{x}}_{k|k}$, and $ \textbf{P}_{k|k}$ are respectively the KF gain, posteriori state estimate and its state covariance matrix.	
	In implementing the KF of \eqref{eq:predict} and \eqref{eq:update}, the variance of the process noise/signal noise of each local KF was informed by our knowledge of the physics of both sensors (electronic perception methods, range resolutions and examining each sensor's depth map to understand the data available to the filter), engineering judgment, and kinematics of the process model. We found these values sufficiently modeled the underlying process dynamics
	\begin{align}
	\sigma_a &= 2000 mm^2; \qquad  {\sigma_{r1}}^2 = 70mm^2 \text{for the Xbox, and} \nonumber \\
	{\sigma_{r2}}^2 &= 60mm^2  \qquad \text{for the Kinect v2 sensor}. \nonumber
	\end{align}	
	\begin{figure}[tb]
		\centering
		\includegraphics[trim=1cm 0cm 4cm 1cm,width=3.6in, clip=true, angle=0, center]{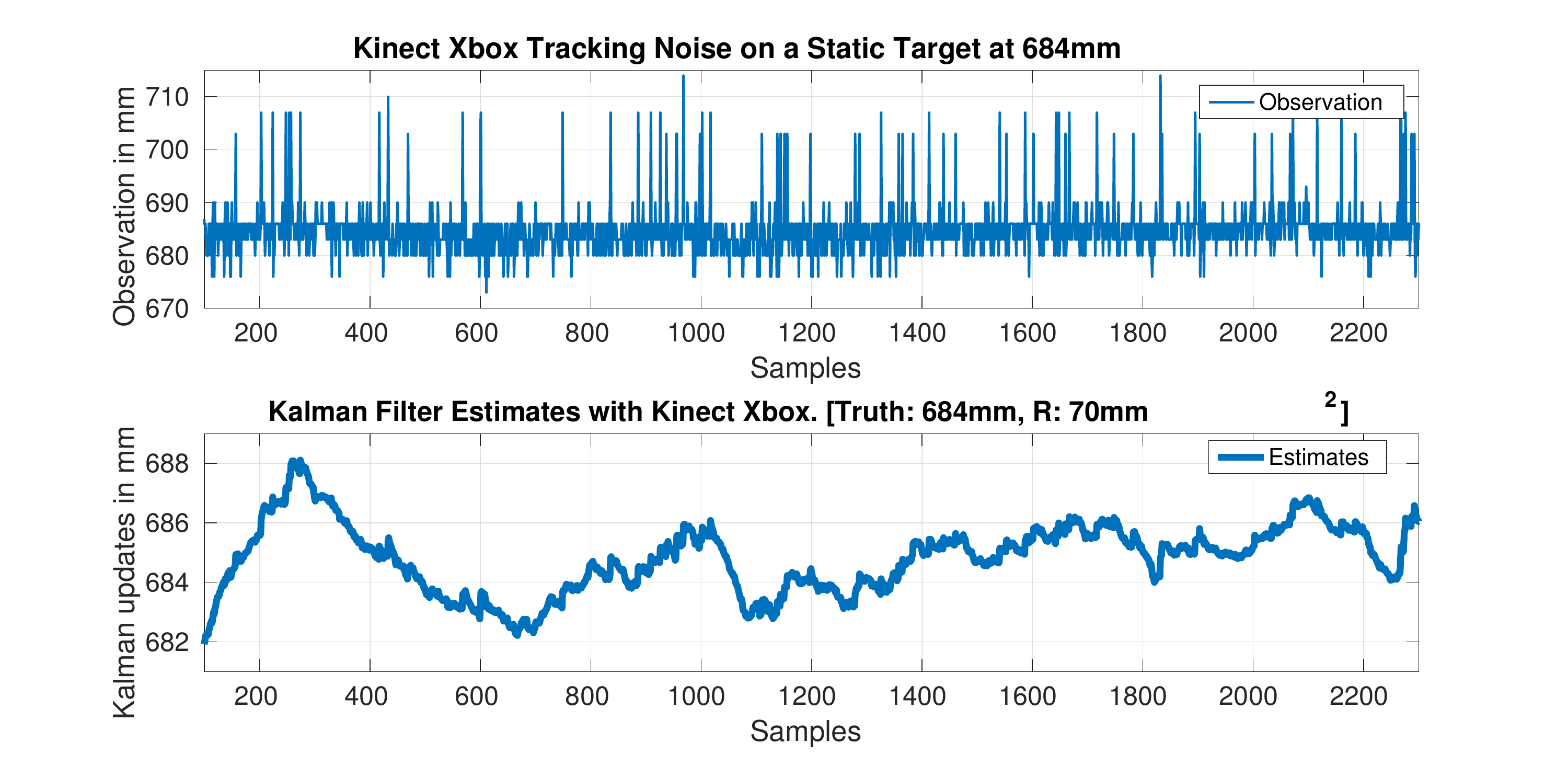}
		%\vspace{0.1em}
		\caption{ KF results for the Xbox observation}
		\label{fig:KFXbox}
		\vspace{-1.6em} 
	\end{figure}	
	\begin{figure}[tb]
		\centering
		\includegraphics[ trim=1cm 0cm 4cm 1cm,width=3.2in, clip=true, angle=0, center]{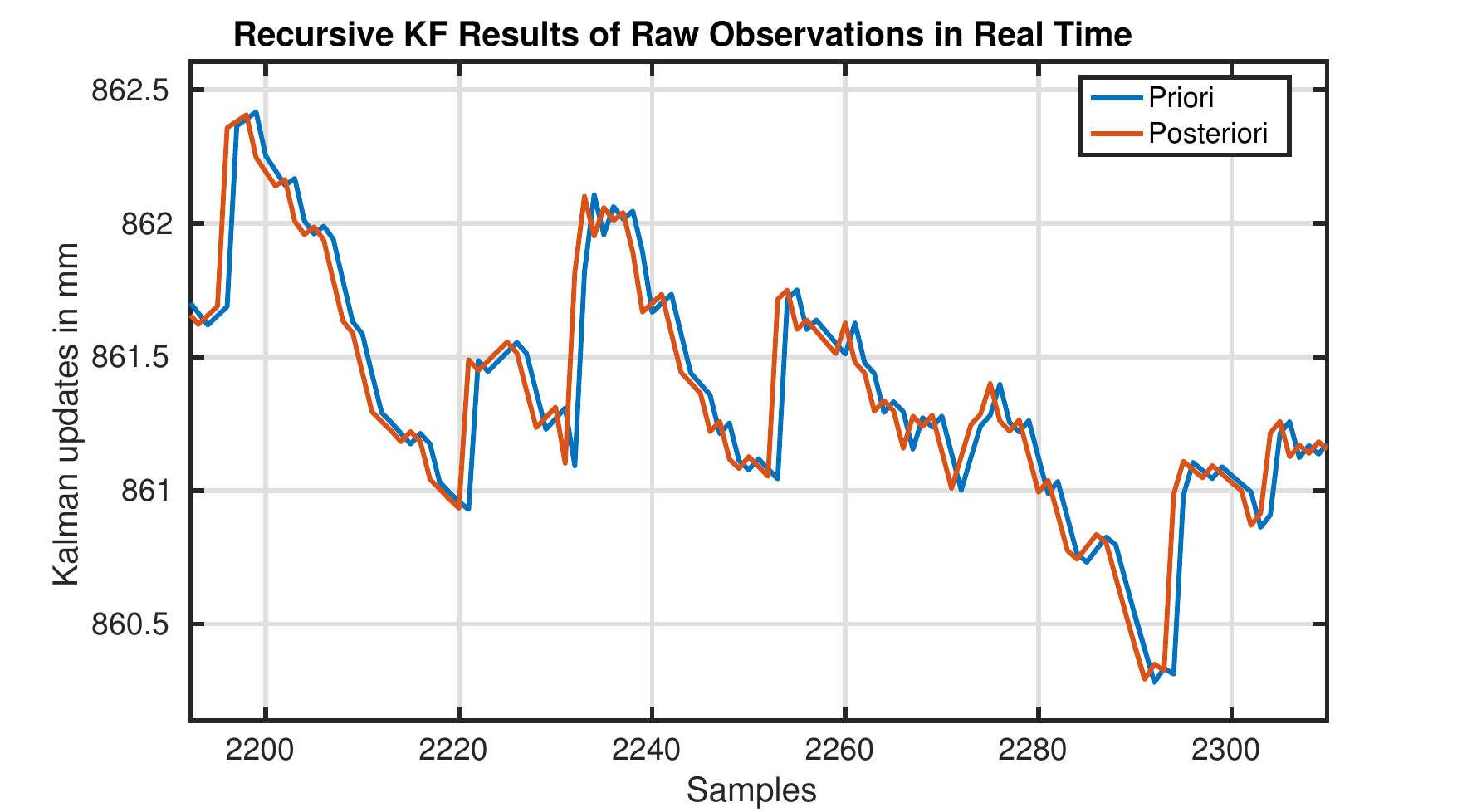} %Kinect2KF.pdf} 
		%\vspace{0.1em}
		\caption{ KF results of Kinect v2's observation}
		\label{fig:KFv2} 
		\vspace{-0.5em}
	\end{figure}	
	Figs. \ref{fig:KFXbox} and  \ref{fig:KFv2} show the local filter estimate results of the observation from both the Kinect Xbox and v2 sensors post-filtering. The noise floor becomes noticeably reduced by each sensor after the KF filtering. 
%	The table in \ref{table:KFXboxv2} describes the results of the steady-state performance of both sensors. 
	The steady-state performance of both sensors include a reduction in the variance of the observation sequence by 80.81\%, while the Kinect v2 shows an improvement in noise rejection by almost 60\% . 
%	\begin{savenotes}
%		\begin{table}[tbp]
%			\centering
%			\caption{Local KF Parameters for Sensors 1 and 2.}
%			\begin{tabular}{|c|c|c|c|c|c|c|}
				%\footnote[1]{Final Prediction Error(Sections 7.4 and 16.4 in \cite{c22})}
				% \footnote[2]{Mean Squared Error ($mm^2$)}
%				\hline \rule[-2ex]{0pt}{5.5ex} 			  & \textbf{Observation Cov.} & \textbf{KF Posteriori Cov.} \\%& \textbf{Observation SNR} & \textbf{KF SNR}\\
%				\hline \rule[-2ex]{0pt}{5.5ex} \textbf{Xbox 360}   & 21.5022    	 & 4.1265 \\	%& -16.0230 dB & 5.2216 \\ 
%				\hline \rule[-2ex]{0pt}{5.5ex} \textbf{Kinect v2}  & 11.4707		 & 4.6325 \\ %&  -4.2967 dB &     \\ 
%				\hline  
%			\end{tabular}
%			\label{table:KFXboxv2}
%		\end{table}
%	\end{savenotes}
	%\begin{tabular}{|c|c|c|c|} 
	%\hline     			& Observation Covariance & KF Estimate Cov. & SNR \\
	%\hline  Kinect Xbox	&  &  &  \\ 
	%\hline  Kinect v2   &  &  &  \\ 
	%\hline 
	%\end{tabular} \label{table:KFXboxv2}
	\subsection{Data Fusion} \label{sec:fusion}	
	Each local KF estimate was combined at a central fusion site to obtain a track-to-track fused global estimate. To communicate each estimate and associated covariance matrix, we create Unix FIFO special files (\textit{i.e. named pipes}) on the kernel file system, write the estimates and covariance matrices to the pipes at each local site and retrieve the values at the central site. 
	
	Named pipes are low-level file I/O systems that can be shared by processes with different ancestry. During data exchange through a FIFO, the kernel forwards all data internally without having to write it to the file system. Since they exist within the kernel and the file system is just an entry serving as a reference point for the processes to access the pipe with a file system name, there is practically no delay in data communication.
	
\begin{figure}[tb]
		\centering
		\includegraphics[trim=1cm 0cm 2cm 0cm, width=3.6in, clip=true, angle=0, center]{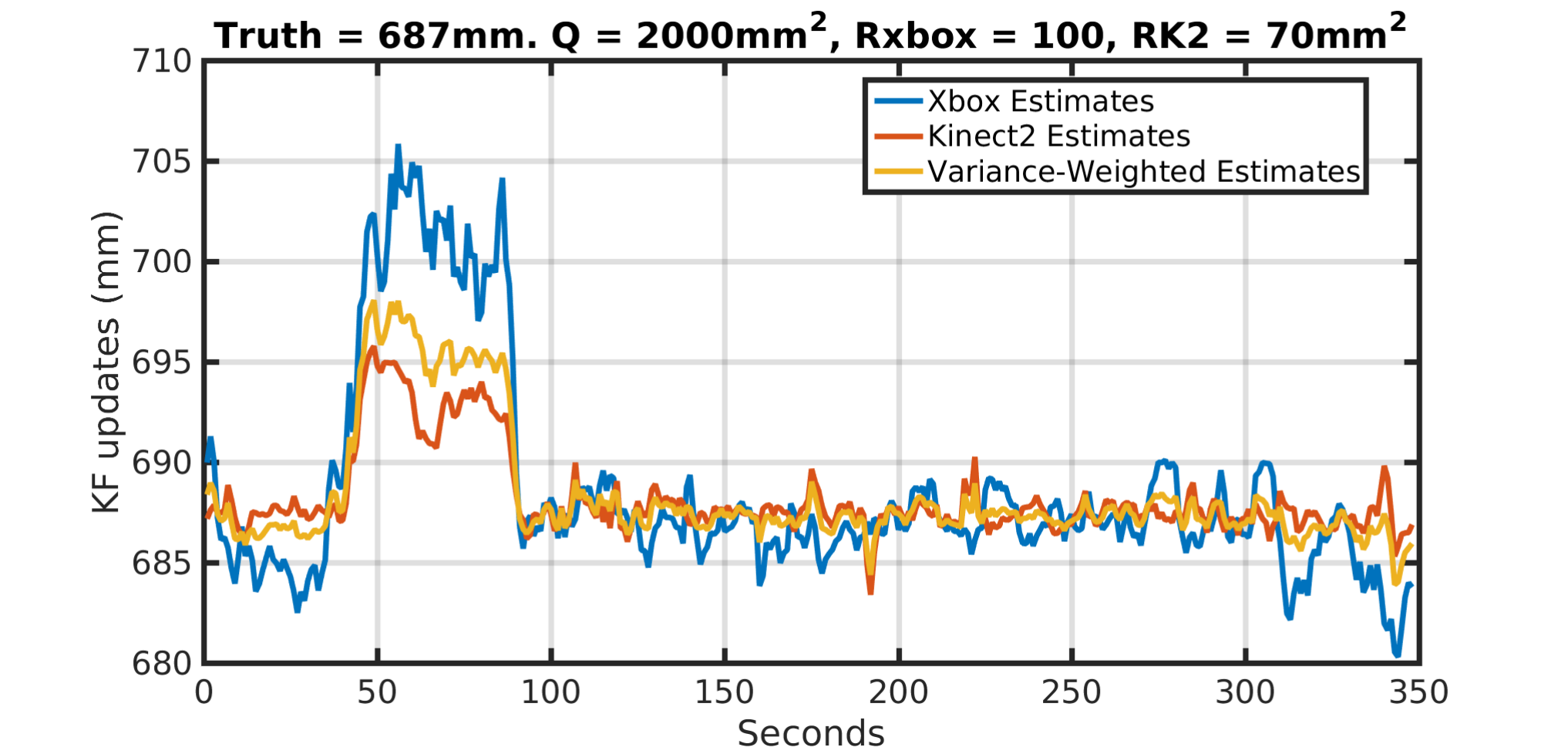}
		%\vspace{0.1em}
		\caption{Kalman filter Track-to-Track fusion of Kinect Xbox and v2's local tracks}
		\label{fig:fusion2} 
	\end{figure}

	Local tracks are generated at each sensor site according to \eqref{eq:predict}, resulting in two local state predictions from the Kalman filters \eqref{eq:state_model}. %, with inputs from the two sensors using independent observation noises, as described in \eqref{eq:sensors_obs}.   
	At the central fusion site, we assume a state model common to both sensors given by \eqref{eq:update} and adopt a variance-weighted average of each local track in the global track fusion algorithm \cite{soft_robot:HDWNotes}
	\begin{align}
	\hat{\textbf{x}}_{F}(k|k) &= \textbf{P}_{F}(k|k)\sum\limits_{i=1}^{N}\left[{\textbf{P}_s}^{-1}(k|k)\hat{\textbf{x}}_s(k|k)\right] \nonumber \\
	\text{where } \textbf{P}_{F}(k|k) &= \left[\sum\limits_{i=1}^{N} {\textbf{P}_s}^{-1}(k|k)\right]^{-1}.
	\end{align}	
%	Figures \ref{fig:fusion1} and 
	\autoref{fig:fusion2} shows the output of the fusion scheme compared against the single Kalman filters during a head-raising motion. The fusion of the local tracks produces better estimates, with improved signal to noise ratio. The fused estimate assigns more weight to the less noisy signal from Kinect v2. Through the implementation of the local tracks and a global track KF estimator, we improved the accuracy of the effective signal to be used in our control algorithm to no more than a standard deviation of 0.75mm from the true position of an object. 
	The noise spikes in the fused tracks when the process state estimates are yet to converge as noticeable in \autoref{fig:fusion2} can be attributed to the noisy initialization of pixels in the sensors  before they attain their steady state values. On average, it takes approximately 30 seconds for the pixel values in the Kinect sensor to reach their final steady state values \cite{KinectDenmark}. This can be avoided by running the fusion algorithm for at least 2 minutes before the fused signal is used for any control purposes.
%	\begin{figure}[tb]
%		\centering
%		\includegraphics[trim=1cm 0cm 2cm 0cm, width=3.6in, clip=true, angle=0, center]{Charts/fusion1.eps}
		%\vspace{0.1em}
%		\caption{ Kalman filter Track-to-Track fusion of Kinect Xbox and v2's local tracks}
%		\vspace{-0.2em}
%		\label{fig:fusion1} 
%	\end{figure} 
	The code for the multisensor fusion experiment is available on the git repos \cite{Xbox_tracker} and \cite{lakehanneKinect2}.

	\section{Soft Robot System Identification} \label{sec:sys_ID}
	We approach the modeling procedure with an identification prediction error (PEM) approach, where we estimate a mathematical model, $G(t)$, based on the minimization of the sum of squared errors between estimates of the head height, $\hat{y(t)}$, and true head height, $y(t)$, from the fusion \emph{i.e.} 
	\begin{equation}
	G(t)= \text{arg } \min_{\theta} \, {V_N(\theta, Z^N)} \nonumber \\
	\end{equation}
	\begin{align} \label{eq:LScrit}
	\text{	where }V_N (\theta, Z^N) = \sum_{k = 1}^{\mathcal{K}}\sum_{i = 1}^{n}\dfrac{1}{2}(\hat{y_i}(k) - y_i(k))^2.
	\end{align}
	$Z^N = \{u(1) \cdots u(N) \quad y(1) \cdots y(N)\}$ is the vector of past input and output (fused estimates) measurements over a bounded interval $[1, N] $ and $\theta$ is the greedy vector of parameters that approximate the model we seek to build. \eqref{eq:LScrit} is a special case of the least squares criterion.
	
	\subsection{Model Structure}
	Following Ljung's formulation in \cite[\S4.5]{soft_robot:LlungBook}, we pose the identification problem as determining the ``best model" from a set of candidate model sets via an iterative approach that parametrizes the noncountable model sets smoothly over an area with the assumption that the underlying system is linear time-invariant. Here, our model structure is a differentiable mapping from a connected, compact subset $\mathcal{D}_{\mathfrak{M}}$ of $\mathcal{R}^d$ to a model set $\mathfrak{M}^*$, such that the gradients of the predictor functions are stable. This procedure is included in the MATLAB system identification toolbox, and since the method is well-documented in \cite{soft_robot:sysidmatlab} we omit details.
	
	External disturbances and stochastic variables are modeled as additive white noise sequence, $e(k)$, based on lagged inputs and outputs, and our objective is to estimate a stochastic state space model structure of the form 
	\begin{align} \label{eq:sysid_stochss}
	\mathbf{x}(k+1) = \mathbf{A x} (k) + \mathbf{B u}(k) + \mathbf{w}(k) \nonumber \\
	\mathbf{y}(k) = \mathbf{Cx}(k) + \mathbf{Du}(k) + \mathbf{v}(k)
	\end{align}
	where the noise terms $\mathbf{w}(k) \text{ and } \mathbf{v}(k)$ compensate for the effect of disturbances beyond frequencies of interest to system dynamics and make the model robust to model uncertainties. Since $u$ and $y$ alone are measurable in our setup, the states $\mathbf{x}(k)$ are estimated and \eqref{eq:sysid_stochss} becomes a linear regression problem, where all the unknown matrix entries are linear combinations of the measured inputs and output variables. This can be written as
	\begin{align} \label{eq:sys_idstate_minimal}
	Y(k) = \Theta \Phi(k) + E(k)
	\end{align}
	where 
	\[ Y(k) = \left[ \begin{array}{c}
	\mathbf{x}(k+1) \\
	\mathbf{y}(k)
	\end{array} \right],
	\hspace{0.3em}
	\Theta= \left[ \begin{array}{cc}
	\mathbf{A} & \mathbf{B} \\
	\mathbf{C} & \mathbf{D}
	\end{array}\right]
	\]
	
	\[\Phi(k) = \left[\begin{array}{c}
	\mathbf{x}(k) \\ \mathbf{u}(k)
	\end{array} \right] \hspace{0.2em} \text{ and  }
	{E}(k) = \left[ \begin{array}{c}
	\mathbb{E}(w(k)) \\
	\mathbb{E}(v(k))
	\end{array}\right].
	\]
	We assume the noise term is white in order to assure an unbiased model. The parameter estimation problem is then to estimate the $\mathbf{A}, \mathbf{B}, \mathbf{C}, \text{ and } \mathbf{D}$ matrices by the linear least squares regression of \eqref{eq:sys_idstate_minimal} assuming no physical insight into the system (i.e. a black box model). $\mathbb{E}(\mathbf{w}(k))$ and  $\mathbb{E}(\mathbf{v}(k))$ are estimated as a sampled sum of squared errors of the residuals.
		
	\subsection{Parameter Estimation} \label{sec:param_est}
	The input,  $u(k)$, and output signals, $y(k)$, can be characterized by a linear difference equation of the form
	\begin{eqnarray}\label{eq:lineardiff}
	y(k) &=-a_1y(k-1)-\cdots-a_{n_a}y(k-n_a) \nonumber \\
	&  -b_1u(k-1) - \cdots -b_{n_b}u(k-n_b) -e(k) \nonumber \\ &-c_1e(k-1) -c_{n_c}e(k-n_c)
	\end{eqnarray}
	where $e(k)$ describes the equation error as a moving average of white noise, and we assume $e(k)$ has a bias-variance term $\lambda$. We can rearrange \eqref{eq:lineardiff} using the vectors 
	\begin{align}\label{eq:psi}
	\begin{split}
	\psi(k, \theta) &= [-y(k-1) \cdots -y(k-n_a) \,\, u(k-1) \cdots \\
	& \qquad u(k - n_b), \,\, e(k-1, \theta), \cdots ,e(k - n_c, \theta)]^T 
	\end{split}
	\end{align}
	\begin{align}
	\theta = [-a_1, \cdots, -a_{n_a}, -b_1, \cdots, -b_{n_b}, -c_1, \cdots, -c_{n_c}].
	\end{align}
	The adjustable parameters of \eqref{eq:psi} are elements of $\theta$. In our prediction model, it is convenient to write \eqref{eq:lineardiff} as a one-step-ahead predictor of the form
	\begin{align}\label{eq:sysid_TF}
	\hat{y}(k) &= G(q, \theta)u(k) + H(q, \theta)\hat{e}(k) \\
	\text{with}  \quad  G(q, \theta) &= \dfrac{B(q)}{A(q)},\text{	 	 } H(q, \theta) = \dfrac{C(q)}{A(q)} \nonumber
	\end{align}
	which is a complete autoregressive moving average with exogenous input (ARMAX) model. $ G(q, \theta)$ represents the transfer function from input to output predictions, and $H(q, \theta)$ denotes the transfer function of prediction errors to the output model, $\hat{y}(k)$;
	% with $A(q)y(k)$ as the autoregressive part, $B(q)u(k)$ as the exogenous input and $C(q)e(k)$ as the moving average of the white noise model; 
	$q$ is the z-transform, $z^{-1}$, while $A(q)$, $B(q)$, and $C(q)$ are polynomials defined as 
	\begin{align}\label{eq:sysid_ABC}
	A(q) &= 1 + a_1 q^{-1} + \cdots + a_{n_a}q^{-n_a} , \nonumber \\
	B(q) &= b_1q^{-1}+ \cdots + b_{n_b}q^{n_b},  \nonumber  \\
	C(q) &= 1 + c_1 q^{-1} + \cdots + c_{n_c}q^{-n_c}
	\end{align}
	\cite{BillingsBook}. The predictor turns out to be a linear filter of the form
	\begin{align} \label{eq:sysid_osa}
	\hat{y}(k|\theta) = W_y(q, \theta)y(k) +W_u(q,\theta)u(k) \\
	\text{and  }
	y(k) = G(q,\theta)u(k) + H(q, \theta)[y(k) - \hat{y}(k)] 
	\end{align}
	where $H(q, \theta)$ is the noise model and $\hat{y}(k)$ above can be regarded as the one-step ahead predictor. After rearranging \eqref{eq:sysid_osa},we find that 
	\begin{equation}
	W_y = 1 - H^{-1}(q, \theta) \nonumber \\
	\text{ and  }
	W_u(q, \theta) = G(q, \theta) H^{-1}(q, \theta)
	\end{equation}
	such that the residual errors from \eqref{eq:sysid_osa} become
	\begin{align} \label{eq:sysid:linearfilter}
	e(k) = [y(k) - G(q, \theta)u(k)]H^{-1}(q,\theta).
	\end{align}
We can consider	\eqref{eq:sysid:linearfilter} as passing the prediction errors through a linear filter that allows extra freedom in dealing with non-momentary properties of the prediction errors. Since the model is that of a linear system, \eqref{eq:sysid:linearfilter} satisfies our objective by approximating the prefilter with the choice of the noise model in \eqref{eq:sys_idstate_minimal}.
	
	%\begin{align} \label{eq:sysid_est}
	%\begin{split}
	%\hat{y}(k|\theta) &= B(q)u(k)+[1-A(q)]y(k) + \cdots\\
	%				& \qquad [C(q) - 1][y(k) - \hat{y}(k|\theta)]
	%\end{split}
	%\end{align}
	%where the identification goal is to find the estimates
	%\begin{align}\label{eq:sysid_modelest}
	%\hat{y}(k|\theta) = \psi^T(k)\theta + K \hat{e}(k)
	%\end{align} 
	%based on past inputs and outputs, 
	%\begin{align}
	%Z^N = \{u(1) \cdots u(N) \quad y(1) \cdots y(N)\}
	%\end{align}
	%and over a time interval $1 \leq k \leq N $ \cite[eq. 4]{soft_robot:paper1} where $\psi(k, \theta)$ is as defined in \autoref{eq:psi} and the noise model $\hat{e}(k)$ is assumed to be white otherwise the model \autoref{eq:sysid_modelest} would be biased and would not correctly give us the true estimates of the underlying system. The predictor of \autoref{eq:sysid_est} can be thought of as a linear filter that corresponds to a one-step ahead predictor with $C(q,\theta)$ as the noise model.
	
	The estimation problem is to predict the estimates, $\hat{y}(k|\theta)$ so that the errors, $\varepsilon(t,\theta) = \parallel y(t) - \hat{y}(t|\theta) \parallel_p$ are minimized by the choice of an appropriate p-norm criterion function, such as the mean squared error proposed in \eqref{eq:LScrit}. % The Stone-Weierstrass theorem allows us to find a  continuous function, $\mathbb{F}$, on the bounded, compact input space with interval [a,b] $\in X$ such that for any $\epsilon > 0$, there is a function $f \in F$ such that for all $x \in X$, $\exists \,\, |Fx - fx| < \epsilon$ \cite{NarendraWeierstrass}.   
	
	\subsubsection{Input Signal Design}
	The input signal choice for a system identification experiment will determine a system's operating point and model accuracy. Therefore, the input should be rich enough to excite a system and force it to show properties needed for the model's purpose. For the model to be informative across all the desired frequency range, a periodic, persistently exciting uniform Gaussian White noise (UGWN) signal with clipped amplitudes corresponding to the bandwidth of the valves was designed offline, and its frequency spectrum analyzed to ensure it had as small a crest factor as possible (since the asymptotic properties of the model will be mostly influenced by the spectrum rather than the waveform's time-series shape). % Therefore, we designed periodic, persistently exciting input signals offline and analyzed the spectrum of each input signal to determine which signal achieved the covered the interested system dynamics with as small a crest factor as possible \cite{soft_robot:LlungLectures} since the asymptotic properties of the model will be mostly influenced by the spectrum rather than the waveform's actual input. 	
	Gaussian White Noise signals (GWN) and Pseudo-Random Binary Signals (PRBS) are well-known to achieve virtually any signal spectrum without very narrow pass bands. %Therefore,we determined their time series signals are designed according to 	
	%\begin{align} \label{eq:mls}
	%\begin{split}
	%u(t) &= Rem(A(q)u(t), 2) = Rem(a_1u(t-1) + \cdots \\
	%     &+ a_5u(t -5), 2)
	%\end{split}
	%\end{align}
	%where $Rem(A(q)u(t), 2)$ is the remainder of the ratio $\frac{A(q)u(t)}{2}$. We carry out the modulo-2 expansion of $A(q)$ in \autoref{eq:mls} to obtain $u(t)$, a binary signal which takes 0 or 1. The MLS sequence has a period of $2^5 -1$ with each period made up of $2^{5-1}$ ones and $2^{5-1}-1$ zeros where we have chosen a fifth polynomial order and scale the current that fully opens the inlet valve by pre-multiplying $u(t)$ by 165. 	
	Therefore, pseudo-random uniform white noise sequences were generated using the very-long cycle random number generator algorithm. Given that the probability density function, $f(x)$, of the uniformly distributed uniform white noise is 
	\begin{align}\label{eq:uwgn}
	f(x) &= \frac{1}{2}A  \qquad \text{if $x < |A|$   and} \nonumber \\
	u(x) &= 0  \qquad \text{  if $x > |A|$ }
	\end{align}
	where $A$ is the amplitude. The expected mean, $\mu$, and the expected standard deviation, $\sigma$ of the sequence are \cite{UWGN}
	\begin{align}
		\mu = \mathbb{E}(x) = 0 \text{,} \qquad
		\sigma = \left[\mathbb{E}\{(x-\mu)\}\right]^{1\over 2} = \frac{A}{\sqrt{3}}.
	\end{align}	
%	the amplitude of input current to the inlet valve, was set to 165 mA to avoid an unbound amplitude of the Gaussian white noise, and a standard deviation $\sigma$ of the noise signal was set to $\frac{A}{\sqrt{3}}$. We excited the valve till we acquired enough samples for the model to be informative. 
	\begin{figure} 
		\centering
		\includegraphics[keepaspectratio=true, width = 3.2in, center, clip = true]{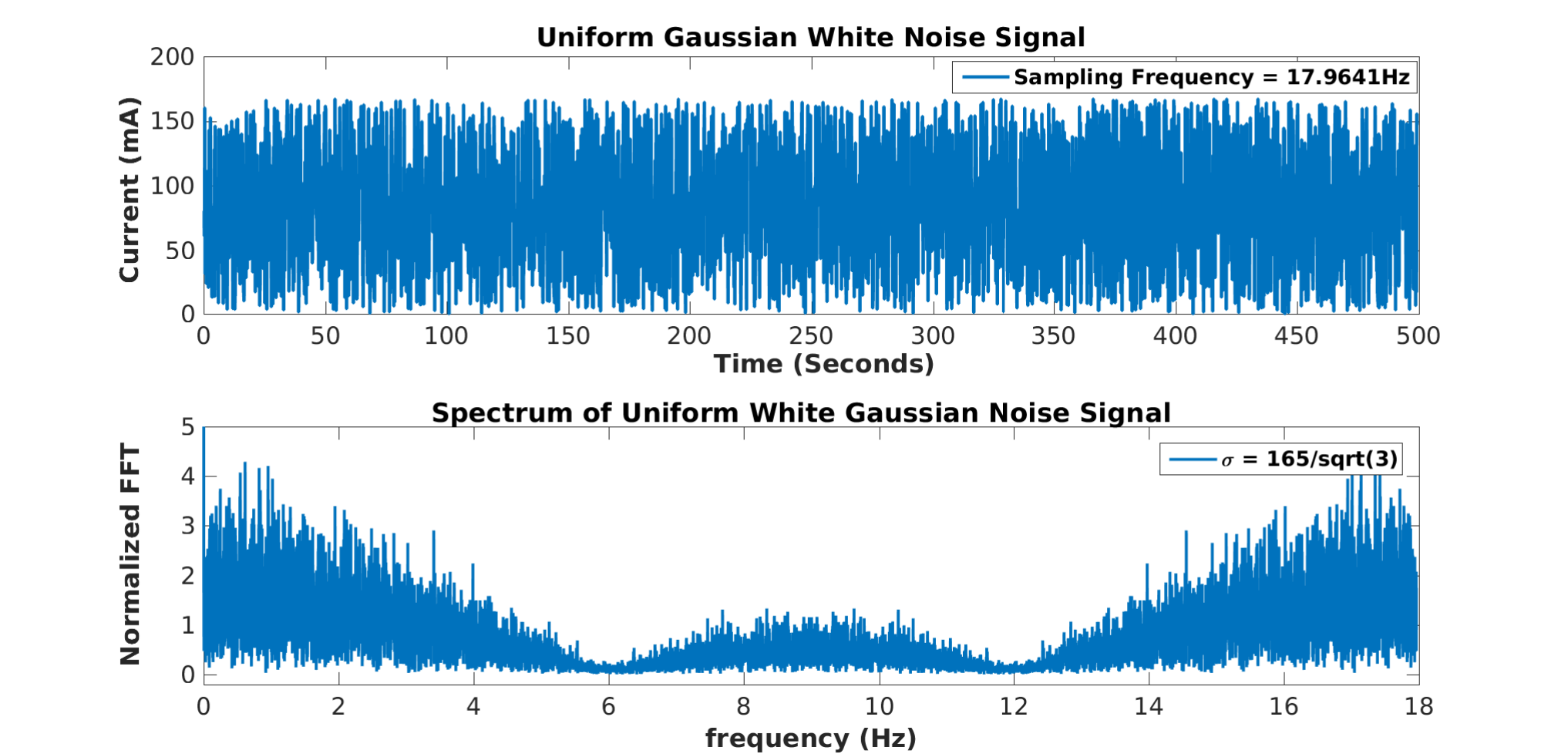}
		\vspace{0.2em}
		\caption{Time/Frequency-Domain Properties of the Input Signal}
		\label{fig:input_spectrum}
		\vspace{-1.2em}
	\end{figure}
	%\begin{align}
	%x_i(k) = \sum\limits_{n=0}^{N-1}\{x_i(n)e^{-j(\frac{2\pi n k}{N})}\}
	%\end{align} 
	The spectrum of the resulting signal in  \autoref{fig:input_spectrum} gives good signal power, which nicely relates to the bandwidth of the pneumatic valves and achieves virtually all signal spectrum with little narrow pass bands. 
	
	We therefore use the signal of \eqref{eq:uwgn} to model the desired asymptotic estimates, $\hat{y}(t)$, of \eqref{eq:sysid_TF}. We sampled $y(t)$, the fused measurement described in \eqref{sec:fusion} well-above the system's Nyquist frequency\iffalse and persistently excited the inlet valve with the uniform white Gaussian noise of \eqref{eq:uwgn}\fi and acquired enough samples to make $Z^N$ asymptotically approach $\hat{\theta}_N$ as $N \rightarrow \infty$. The data collection procedure closely follows that described in our previous paper and we refer readers to \cite[\S IV.A]{soft_robot:paper1} for a more detailed treatment.
	
	The collected data was separated in a 60:40\% ratio for training and testing purposes, respectively, to assure a training model that generalizes well. 
	\subsubsection{State Space Realization}
	\iffalse
	With the assumption that $u(k)$ is piecewise constant over the sampling interval, $K$, sampling $G(q,\theta)$ and $H(q,\theta)$ over $T$ in \eqref{eq:sysid_TF}, the estimates become
	\begin{align}
	x(kT + T) &= A_T(\theta)x(kT) + B_T(\theta)u(kT) + w(kT)  \nonumber \\
	\text{where }\qquad A_T(\theta) &= e^{F(\theta)T}  \nonumber \\
	\text{and   }\qquad B_T(\theta) &= \int\limits_{\tau = 0}^{T} e^{F(\theta)\tau} G(\theta) d\tau,
	\end{align}
	$F \in \mathbb{R}^{n\times n}$ and $G \in \mathbb{R}^{n \times m}$ are the matrices that map the states, $x(kT)$, and input, $u(kT)$, into $x(kT + T)$. And
	\begin{align} \label{eq:sysid_y(t)}
	y(t) = G_T(q, \theta)u(t) + v_T(t), 	\hspace{0.4em} t = T, 2T, 3T, \ldots
	\end{align}
	\fi

	If we define $$\hat{Y}_r(k)= [\hat{y}(k|k-1), \cdots, \hat{y}(k+r-1)| k -1 ]^T$$  $$\hat{Y} = [\hat{Y}_r(1) \cdots \hat{Y}(N)],$$ it follows that 1) as $N \rightarrow \infty$, there are $n$-th order minimal state space descriptions of the system if and only if the rank of the matrix of prediction vectors, $\hat{Y}$, is equal to $n$ for all $r \geq n$; and 2) the state vector of any minimal realization in innovations can be chosen as linear combinations of $\hat{Y}_r$ that form a row basis for $\hat{Y}$, i.e., $$x(t) = L \hat{Y}_r(k)$$ with $L$ being an $n\times pr $ matrix ($p$ is the dimension of $y(k)$) \cite[\S7.3]{soft_robot:LlungBook}. The true prediction is given by \eqref{eq:sysid_TF} with innovations $e(j)$ written as a linear combination of past input-output data. The predictor can thus be expressed as a linear function of $u(i), \, y(i), \, i \le k -1 $. In practice, the predictor is approximated so that it depends on a finite amount past data such as $s_1$ past outputs and $s_2$ past inputs of the form
	\begin{align}
	\hat{y}(k|k-1) &=\alpha_1 y(k-1)+ \cdots +\alpha_{s_1}y(k-s_1) \\ \nonumber
	& + \beta_1 u(k-1)+ \cdots + \beta_{s_2}u(k-s_2).
	\end{align} 	
	Piping the identification data through the MATLAB function \textsf{`ssest'} and testing various model orders based on the ranking of singular values of the Hankel matrix of input-output measurements \cite{soft_robot:LlungBook}, we obtained the results listed in \autoref{table:modelshopping} on training and testing dataset. The MATLAB system identification script is provided on a github repo \cite{BBIdentSRS} and contains the dataset used for the experiment.	
	\begin{savenotes} 	
		\begin{table}[tb]
			\centering
			\caption{Model estimates}
			\label{table:modelshopping}
			\begin{tabular}{|c|c|c|c|c|c|}
				\hline \rule[-2ex]{0pt}{5.5ex} \textbf{Data Type} &\textbf{Expts} &\textbf{MO}\footnote[1]{Model Order} & \textbf{MSE}\footnote[2]{Mean Squared Error ($mm^2$).} & \textbf{Fit (\% )} & \textbf{FPE} \footnote[3]{Akaike Final Prediction Error (\cite[Secs 7.4 and 16.4]{soft_robot:LlungBook}).} \\
				\hline \rule[-2ex]{0pt}{5.5ex} \textbf{Training} & i & 2 & 0.001437 & 97.64 & 0.001438\\ 
				& ii   	& 4 & 0.001454 & 97.62 & 0.00145584\\ 
				& iii  	& 6 & 0.001333 & 97.72 & 0.001336\\
				& iv    & 8 & 0.001298 & 97.76 & 0.001298\\  
				\hline  \rule[-2ex]{0pt}{5.5ex} \textbf{Testing} &i & 2 & 0.000963 & 98.47 & 0.000964\\ 
				& ii    & 4 & 0.0008574 & 98.56 & 0.008594	\\
				& iii   & 6 & 0.000846 & 98.57 & 0.000849\\ 
				& iv    & 8 & 0.000843 & 98.57 & 0.000848\\ 
				\hline %\rule[-2ex]{0pt}{5.5ex} %& & & & 
			\end{tabular}
			\vspace{-0.8em}
		\end{table}
	\end{savenotes}	
	The model set above exhibit a high-fit of estimate to fed data with generally good mean-square errors and final prediction errors for a control experiment. With increasing model order starting from 4, we see that the fits start reaching convergence, as the mean-square errors and final prediction errors become constant. In the frequency-domain, this is the equivalent to  having pole-zero cancellations for higher-order models. We therefore conclude there is no useful properties a higher-order model could predict beyond an order of 8. The second-order model sufficiently approximates the system and is not significantly outperformed by the higher order models --which would contribute higher complexity to the control design. \iffalse There is also little to gain in choosing a model order of 4 over a 2nd-order based on the estimation results presented in the table. A 4th order state space model would mean more complexity in implementation of control design. Besides, a good controller should be able to tolerate model uncertainties and system properties not captured by our model.\fi We therefore pick the 2nd order state space model \eqref{eq:sysid_stochss} as
	\begin{align} \label{eq:statemodel} 	
	\textbf{x}(k+Ts) = \textbf{A} \textbf{x}(k) + \textbf{B} \textbf{u}(k) + \textbf{K} \textbf{e}(k) \nonumber \\
	\textbf{y}(k) = \textbf{C} \textbf{x}(k) + \textbf{D} \textbf{u}(k) + \textbf{e}(k)
	\end{align} 	
	where $Ts$ is the sampling period, $\textbf{e}(k)$ is the modeled zero-mean Gaussian white noise with non-zero variance, 
	\iffalse
	\begin{align}
	\mathbb{E}[Ke(k)\,Ke'(\tau)]&=Q(k)\delta(k-\tau) \qquad \mathbb{E}[Ke(k)] &\equiv 0 \nonumber \\
	\mathbb{E}[e(k)e'(\tau)] &=R(k)\delta(k-\tau) \qquad \mathbb{E}[e(k)] &\equiv 0 
	\end{align}
	\fi
	\begin{align}  \label{eq:ControlSSModel}
	\textbf{A} = 
	\begin{bmatrix}
	0    &    1	\\
	-0.9883 &   1.988
	\end{bmatrix}, \hspace{0.4em}
	\textbf{B} =
	\begin{bmatrix}
	-3.03e-07 \\
	-4.254e-07
	\end{bmatrix} \nonumber \\
	\textbf{C} = 
	\begin{bmatrix}
	1  &  0
	\end{bmatrix}, \hspace{0.4em} D = 0, \text{and} \hspace{0.4em} \textbf{K} = 
	\begin{bmatrix}
	0.9253 &  0.9604
	\end{bmatrix}^T.
	\end{align}
The \iffalse resulting state-space model obtained in \eqref{eq:statemodel} is in the observability-canonical form; the\fi pair $(A, B)$ is stabilizable and and the pair $(A, C)$ is detectable. 
	
	\section{LQG Control} \label{sec:LQG}
	\iffalse
	Given the linear time-invariant system of \eqref{eq:ControlSSModel}, our goal is to find a control input history based on a receding horizon that minimizes a well-posed \footnote{A well-posed cost function should have unbiased predictions at steady state.} cost function with respect to a predictive control law over a defined prediction horizon, such that the minimization problem is consistent with offset-free tracking.  Since the algorithm is implemented on a computer, we design a full-state observer in discrete time that estimates the system states identified in \autoref{sec:param_est}.
	\fi
	 We employ a LQG controller and estimator to minimize the following cost function subject to the state equation \eqref{eq:ControlSSModel}	
	\begin{equation}  \label{eqn:LQ-cost}
	J = \sum\limits_{k=0}^{\mathcal{K}} x^T(k)\,Q\,x(k) + R \, u(k)^T \, u(k) + 2 x(k)^T \, N \, u(k)
	\end{equation}  
	where $\mathcal{K}$ is the terminal sampling instant, $Q$ is a symmetric, positive semi-definite matrix that weights the $n$-states of the $A$ matrix, $N$ specifies a matrix of appropriate dimensions that penalizes the cross-product between the input and state vectors, while $R$ is a symmetric, positive definite weighting matrix on the control vector $u$. The quadratic cost function in \eqref{eqn:LQ-cost} allows us to find an analytical solution (controller sequence) to the minimization of $J$ over the prediction horizon, $n_y$	
	\begin{equation} \label{eqn:min J} 
	\Delta u  = \text{arg } \underset{\Delta u}{\text{min }}J 
	\end{equation}
	where $\Delta{u}$ is the future control sequence and the first element in the sequence is used in the control law at every time instant.	
	We model additive white noise disturbances into the discrete estimator's states; therefore the optimization problem becomes a stochastic optimization problem that must be solved. 
	
The separation theorem ensures that we can construct a state estimator which asymptotically tracks the internal states from observed outputs, $y(k)$, using the algebraic Riccati equation given as 
	\begin{equation}    \label{eqn:Riccati}
	\begin{split}
	A^T P A \mbox{-}(A^T P B \mbox{+} N)(R \mbox{+} B^T P B)^{-1}(B^T P A \mbox{+} N) 
	\mbox{+} & Q.
	\end{split}
	\end{equation}
	where $P$ is an unknown $n \times n$ symmetric matrix and $A$, $B$, $Q$, and $R$ are known coefficient matrices as in \eqref{eq:ControlSSModel} and \eqref{eqn:LQ-cost}. We find an optimal control law by solving the minimization of the LQ problem, \eqref{eqn:LQ-cost} which we then feed into the states. 
	
	In practice, it is a good idea to start with an identity matrix, $Q$, a zero penalty matrix, $N$, and tune $R$ till one obtains convergence by the state estimator. The following optimal values were used after a heuristic search
	\begin{align}
	Q = 
	\begin{bmatrix}
	1.0566 & 0 \\
	0      & 1.0566 
	\end{bmatrix},  \hspace{0.4em} 
	R = 
	\begin{bmatrix}
	0.058006
	\end{bmatrix}.
	\end{align}

	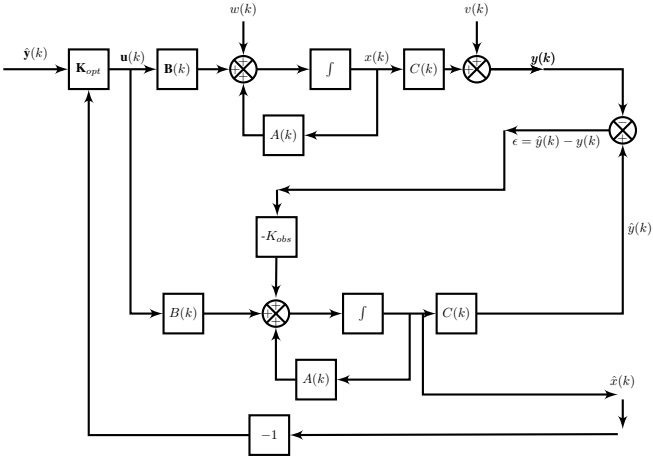
\begin{figure} 
		\centering
		\begin{tikzpicture}[thick,scale=0.5, every node/.style={transform shape}]
		\sbEntree{E}
		\sbBloc[4]{bloc1}{$\textbf{B}(k)$}{E}
		\sbRelier[$\textbf{u}(k)$]{E}{bloc1}
		\sbBloc[-2.7]{controller}{$\textbf{K}_{opt}$}{E}
		
		\sbSortie[-8]{starto}{controller}
		\sbRelier{starto}{controller}
		\sbNomLien[0.8]{starto-controller}{$\hat{\textbf{y}}(k)$}
		
		%\sbDecaleNoeudy[1.2]{controller}{controlpt2}	\sbSortie[-4.5]{starto2}{controlpt2}
		%	\sbRelier{starto2}{controlpt2}
		%	\sbNomLien[0.8]{starto2-controlpt2}{}
		
		\sbCompSum[5]{sumbloc}{bloc1}{+}{+}{+}{}
		\sbRelier{bloc1}{sumbloc}
		%\sbBlocL{Summand}{$\Sigma$}{bloc1}
		\sbBloc[4]{integral}{$\int$}{sumbloc}	
		\sbRelier{sumbloc}{integral}
		\sbBloc[4]{Hprime}{$C(k)$}{integral}
		\sbRelier[$x(k)$]{integral}{Hprime}
		\sbCompSum[4]{sumtwobloc}{Hprime}{+}{}{+}{}
		\sbRelier{Hprime}{sumtwobloc}
		\sbDecaleNoeudy[-4]{sumbloc}{u}
		\sbDecaleNoeudy{integral}{v}
		\sbBlocr{F}{$A(k)$}{v}
		\sbRelieryx{integral-Hprime}{F}
		\sbRelierxy{F}{sumbloc}
		\sbRelier{u}{sumbloc}   	
		\sbNomLien[0.5]{u}{$w(k)$}
		
		\sbSortie[4]{y}{sumtwobloc}
		\sbRelier{sumtwobloc}{y}
		\sbNomLien[0.8]{y}{$y(k)$}
		
		\sbDecaleNoeudy[-4]{sumtwobloc}{w}	
		\sbRelier{w}{sumtwobloc}
		\sbNomLien[0.5]{w}{$v(k)$}
		
		\sbDecaleNoeudy{sumtwobloc-y}{v2}
		\sbCompSum[8]{sum3bloc}{v2}{-}{+}{}{}
		\sbRelierxy{y}{sum3bloc}
		% % Estimator circuit
		\sbDecaleNoeudy[9]{bloc1}{Ge}
		\sbDecaleNoeudy[9.5]{Ge}{Gee}
		\sbBlocr[-2]{Ghat}{$B(k)$}{Gee}
		
		\sbSortie[2]{GhatGee}{E}
		\sbRelieryx{GhatGee}{Ghat}
		
		%\sbRelierxy{E}{Ghat}
		\sbCompSum[7]{sume1}{Ghat}{+}{+}{+}{}
		\sbRelier{Ghat}{sume1}
		\sbBloc[4]{integrale}{$\int$}{sume1}	
		\sbRelier{sume1}{integrale}
		\sbBloc[4]{Hprimee}{$C(k)$}{integrale}
		\sbRelier{integrale}{Hprimee}
		%	\sbNomLien[3]{Hprimee}{$\hat{y}(k)$}
		
		\sbRelierxy[$\hat{y}(k)$]{Hprimee}{sum3bloc}
		\sbDecaleNoeudy[-4]{sume1}{ue}
		\sbDecaleNoeudy{integrale}{ve}
		\sbBlocr{Fe}{$A(k)$}{ve}
		\sbRelieryx{integrale-Hprimee}{Fe}
		\sbRelierxy{Fe}{sume1}	
		%  	\sbNomLien[0.5]{u}{$v(k)$}
		\sbDecaleNoeudy[9]{sum3bloc}{xe}
		\sbDecaleNoeudy[11]{xe}{xee}
		
		\sbDecaleNoeudy[-5.8]{sume1}{Ke}
		\sbBloc[-1.5]{sume1}{-$K_{obs}$}{Ke}
		\sbDecaleNoeudy{Ke}{sumpt}
		\sbRelier{sume1}{sumpt}
		
		\sbSortie[-10]{arr}{sum3bloc}
		\sbRelier[$\epsilon = \hat{y}(k) - y(k)$]{sum3bloc}{arr}  	
		%\sbLien{sum3bloc}{arr} 
		\sbDecaleNoeudy[9.5]{sumbloc-integral}{wL}
		\sbSortie[-0.5]{wLL}{wL}	
		\sbRelieryx{arr}{wLL}    
		\sbSortie[-1.5]{Ltop}{sume1}
		\sbDecaleNoeudy[-1.2]{Ltop}{Ltopey}
		\sbRelier{wLL}{Ltopey}
		%\sbNomLien[0.5]{w}{$w(k)$}
		
		\sbSortie[3]{we}{integrale} 
		\sbRelieryx{we}{xee}
		\sbNomLien[1]{xee}{$\hat{x}(k)$}
		
		\sbDecaleNoeudy[3]{xee}{xeebutt}
		\sbRelier{xee}{xeebutt}
		\sbDecaleNoeudy[4.2]{Fe}{Ae}
		\sbBlocr{ngtv}{$-1$}{Ae}
		\sbRelier{xeebutt}{ngtv}
		\sbRelierxy{ngtv}{controller}
		
		%\sbRelierxy{xee}{controller}
		%
		\sbSortie[4]{y}{sumtwobloc}
		\sbRelier{sumtwobloc}{y}
		\sbNomLien[0.8]{y}{$y(k)$}    
		\end{tikzpicture}
		\caption{Full Linear Quadratic Gaussian Plant Estimator} 
		\label{fig:lqg}
		\vspace{-1.4em}
	\end{figure}
		
	We construct a full online estimator for the identified plant as in \autoref{fig:lqg}, whereby the noise processes are assumed to be independent, white, Gaussian,  of zero mean and known covariances. The optimal controller gains, $K_{opt}$, are determined from the equation
	\begin{equation}
	K_{opt} = {R}^{-1}(B^T \, P + N^T)
	\end{equation}
	\cite{Anderson} where $P$ is the solution to the algebraic Riccati equation \eqref{eqn:Riccati} and 
	$
	\mathbb{E}[w(k)w'(\tau)] = R(k) \delta(k-\tau).
	$	
	Therefore, the online optimal estimate, $\hat{x}(k+1)$ of $x(k)$ is  
	\begin{align}
	\hat{x}(k+1) = A(k)\hat{x}(k) + K_{lqg}\left[C(k)\hat{x}(k)-y(k)\right]
	\end{align}
	where $\hat{x}(k_0)=\mathbb{E}\left[x(k_0)\right]$
	The observer is equivalent to a discrete stochastic Kalman filter that estimates the optimal state $\hat{x}(k|k)$ as shown in \autoref{fig:lqg}. The estimator equations are similar to equations \eqref{eq:predict} and \eqref{eq:update} and the online, unbiased estimate is	
	\begin{align}
	\hat{x}(k+1) &= A(k)\hat{x}(k) - K_{obs}[\hat{y}(k)- y(k)] + B(k)u(k) \nonumber \\
	\hat{y}(k) &= C(k)\hat{x}(k)
	\end{align}
	$\implies$
	\begin{align}
	\begin{split}
	\hat{x}(k+1) &= A(k)\hat{x}(k) - K_{obs}[C(k)\hat{x}(k)- y(k)]\\
	& \qquad \qquad \qquad + B(k)u(k).
	\end{split}
	\end{align}
%	Using the plant equation of \autoref{eq:sysid_stochss}, we find that
%	\begin{align} \label{eq:state_error}
%	\begin{split}
%	\dfrac{d}{dk}[x(k) - \hat{x}(k)] &= \hat{A}[x(k) - \hat{x}(k)] + \\& \qquad v(k) + K_{lqg}(k)w(k)
%	\end{split}
%	\end{align}
%	where the error equation for $x(k) - \hat{x}(k)$ is driven by zero mean process, $v(k)$ and $w(k)$ and $\mathcal{E}(x(k) - \hat{x}(k) = 0.$ Therefore the estimate of $x(k)$ is unbiased.	
	Through heuristics, we found the following variances of the online estimator to be useful:
	\begin{align}
	Q_e = \begin{bmatrix}
	0.4511    &    0	\\
	0         &   0.4511
	\end{bmatrix}, \hspace{0.4em}
	R_e =
	\begin{bmatrix}
	0.01 
	\end{bmatrix} \nonumber 
	\end{align}
	
	\section{Experimental Results and Discussion} \label{sec:exps}
	The control network was implemented on an NI-myRIO running LabVIEW 2015. We initialized the Kinect sensors to allow for all pixels within the depth cameras to reach steady state under ambient light. We performed multiple experiments to evaluate the developed state space model of \ref{sec:param_est} and LQG controller of \ref{sec:LQG}. \footnote{The LabVIEW identification and control codes are available on the git repo \cite{LQGDesign}.}. The input variable is the current that excites the valve, which in turn actuates the bladder; the head moves in response to bladder actuation. The fused estimate of the Kinect sensors are used to estimate the real-time head pitch motion as described in \ref{sec:fusion}; this is in turn used in a feedback to the LQG controller. 
	
\begin{figure}[tb]
		\centering
		\includegraphics[keepaspectratio = true, width=3.6in, clip=true, angle=0, center]{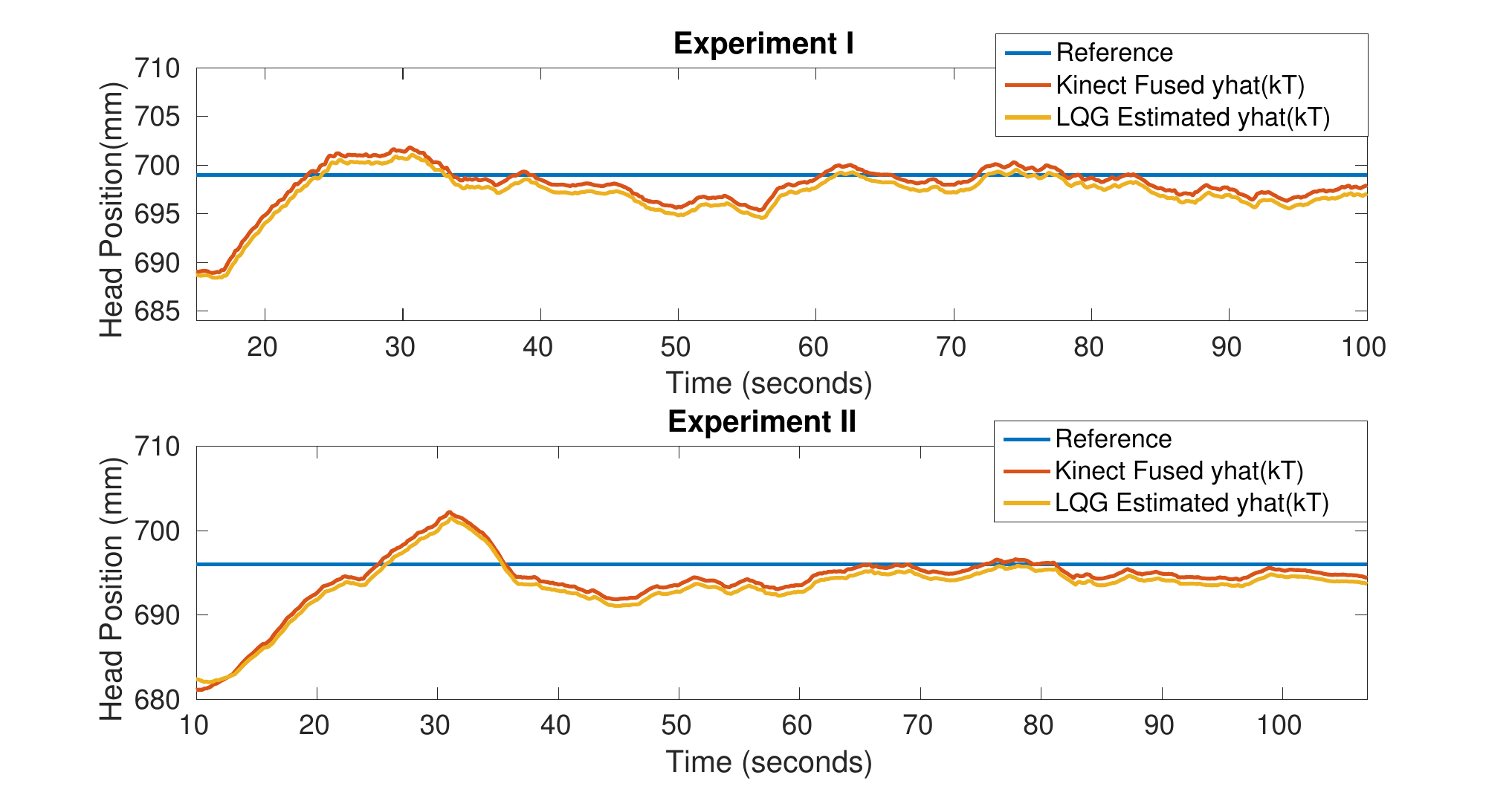}
		%\vspace{0.1em}
		\caption{ LQG Controller on Manikin Head}
		%\vspace{-1.8em}
		\label{fig:LQGI} 
	\end{figure}

	\autoref{fig:LQGI} shows the results from a constant reference trajectory, which the head is meant to track.  We notice a settling time of approximately 24 seconds before we reach steady state. The delay arises from our design requirements and is not a drawback in clinical trajectory tracking where we must ensure smooth head motion to desired target. It is also seen that the controller exhibits relatively  smooth tracking within a 1.5 mm standard deviation over time after a relative overshoot of 5mm in bottom graph of \autoref{fig:LQGI}. The overshoot can be explained by the estimator's search for a steady state region based on the time it takes for the pixel values of the sensors to reach steady state. The controller tracks the reference to within $\pm2mm$.
		
\begin{figure}[tbp]
			\centering
			\includegraphics[keepaspectratio = true, width=3.6in, angle=0 , center]{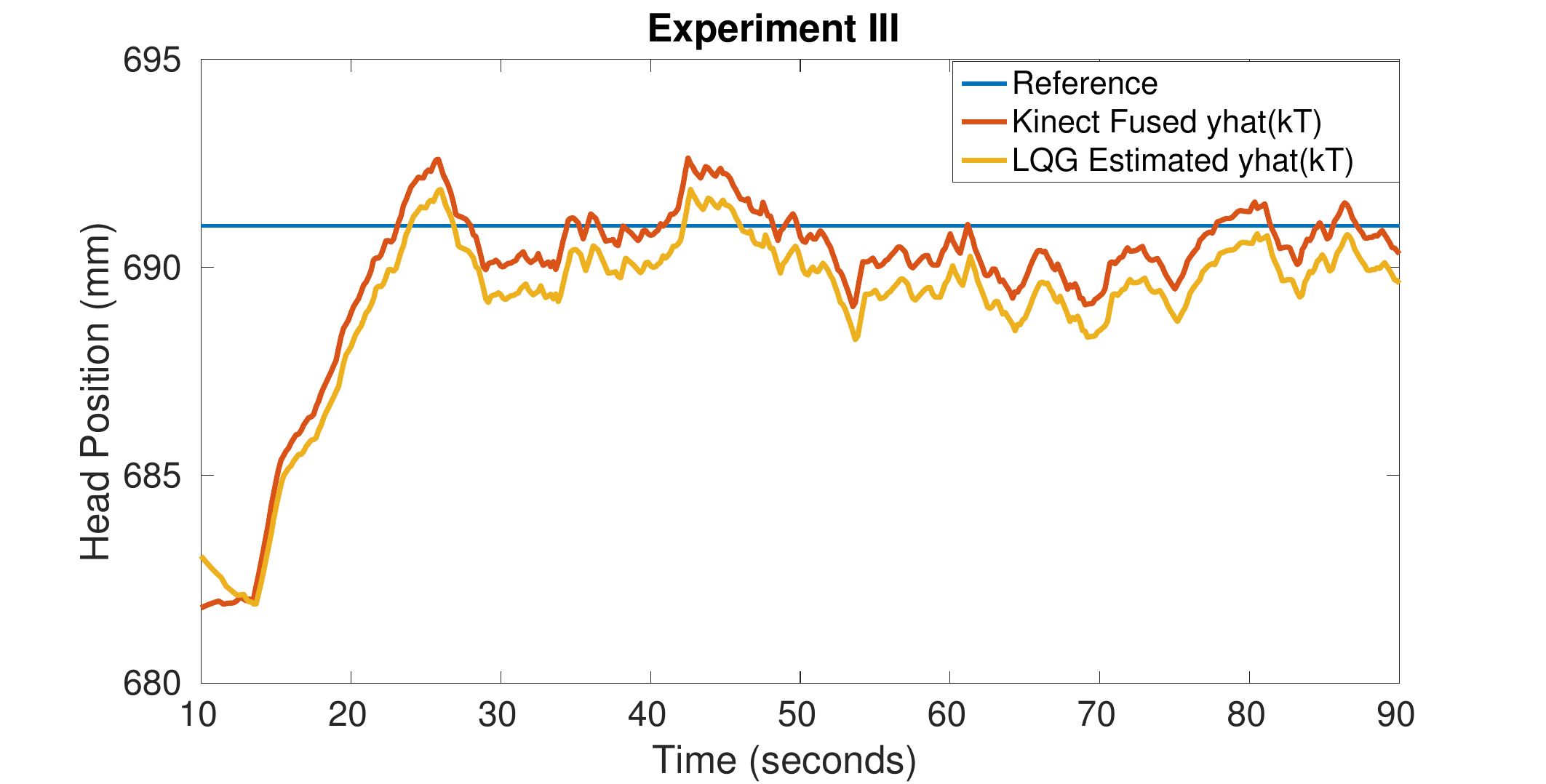}
			\caption{ LQG Controller on Manikin Head}
			\label{fig:LQGII} 
			\vspace{-0.9em}
		\end{figure}

		However, we noticed an inconsistency at certain operating ranges in the current LTI model.  The applied current based on fusion feedback occasionally reaches a steady state error, as can be seen from \autoref{fig:LQGII}. %The inherent settling time of $~10$ seconds of the sensors also affects the fusion algorithm. 
We conjecture this is due to an unmodeled nonlinearity at the inlet  valve that maps input currents to system states. To better approximate the nonlinearity from input to output, we will investigate using a Hammerstein block-structured model %\autoref{fig:Hammerstein} 
that better approximates the nonlinearity from inputs to states and states to output of the system. 
			%\begin{figure}[tb]  
				%\centering
				%\begin{tikzpicture}[thick,scale=0.7, every node/.style={transform shape}]			
				%\sbEntree{E}
				%\sbBloc[4]{gdot}{$g(.)$}{E}
				%\sbRelier[$u$]{E}{gdot}
				%\sbBloc{Hdot}{$G(z^{-1})$}{gdot}
				%\sbRelier[$w$]{gdot}{Hdot}
				%\sbCompSum[5]{sumpt}{Hdot}{}{}{}{}
				%\sbRelier{Hdot}{sumpt}
				%\sbSortie[3]{lastpt}{sumpt}
				%\sbRelier[$y$]{sumpt}{lastpt}
				%\sbDecaleNoeudy[4]{sumpt}{eta}
				%\sbRelier[$\mu$]{eta}{sumpt}
			%\end{tikzpicture}
		%\caption{The Hammerstein model structure}
		%%\vspace{-1.5em}
				%\label{fig:Hammerstein}
		%\end{figure}
		%In the Hammerstein model shown, the $g(.)$ block represents the static nonlinearity that maps inputs to states while the $G(z^{-1})$ block denotes the following maps the linear dynamics of the system's states to the sensors' measurements. We reason that this will minimize the mapping errors from input to output signals and provide better offset-free tracking.
					
	The fusion algorithm proved useful to cancel jitter in the depth measurement of the sensor over our previous results, but it falls short of the 1mm accuracy in head and neck cancer RT. Having established proof of concept in this investigation, we will begin investigation of better head localization by using sophisticated motion capture systems or laser scanners, such as those actively employed in IGRT.	
		% To overcome this, in an ongoing work, we are adopting the Vicon motion capture (`mocap') system for localized tracking of the head. The mocap system consists of ten different infra-red wavelength cameras with 12.5mm lens performing simultaneous localization of a body at frame rates up to 100Hz. We have tested this on the human body animations and the model used in this work and it has pixel-accuracy of 1mm without filtering. 
	
\section{Summary}
	We have presented a continuation of or initial investigation into a pneumatically-driven soft robot system for head and neck radiotherapy.  Measurements from two Kinect RGB-D cameras are fused to refine the accuracy of observations. System identification on the soft robot was combined with LQG controller design to provide	an optimal controller. Experiments showed we could actuate the patient head within  2.5mm accuracy.  Future work will investigate accurate multi-axis  positioning using better 3D sensor, control of multiple IABs, and nonlinear modeling or model free-methods to overcome limitations of the LTI model.
	
	\addtolength{\textheight}{-0.1cm}   % This command serves to balance the column lengths
	% on the last page of the document manually. It shortens	% the textheight of the last page by a suitable amount.
	
	%%%%%%%%%%%%%%%%%%%%%%%%%%%%%%%%%%%%%%%%%%%%%%%%%%%%%%%%%%%%%%%%%%%%%%%%%%%%%%%%

	%%%%%%%%%%%%%%%%%%%%%%%%%%%%%%%%%%%%%%%%%%%%%%%%%%%%%%%%%%%%%%%%%%%%%%%%%%%%%%%%

	%%%%%%%%%%%%%%%%%%%%%%%%%%%%%%%%%%%%%%%%%%%%%%%%%%%%%%%%%%%%%%%%%%%%%%%%%%%%%%%%
	
	%\section*{APPENDIX}

	%%%%%%%%%%%%%%%%%%%%%%%%%%%%%%%%%%%%%%%%%%%%%%%%%%%%%%%%%%%%%%%%%%%%%%%%%%%%%%%%
	\providecommand\BIBentryALTinterwordstretchfactor{2.5}
	\bibliographystyle{IEEEtran}
	\bibliography{soft_robot}	
\end{document}